\documentclass[nohyperref]{article}

\usepackage{microtype}
\usepackage{graphicx}
\usepackage{subfigure}
\usepackage{booktabs} 

\usepackage{hyperref}

\makeatletter
\newcommand\theHALG@line{\thealgorithm.\arabic{ALG@line}}
\makeatother

 \usepackage[accepted]{icml2022}

\usepackage{amsmath}
\usepackage{amssymb}
\usepackage{mathtools}
\usepackage{amsthm}

\usepackage[capitalize,noabbrev]{cleveref}
\theoremstyle{plain}

\theoremstyle{definition}

\theoremstyle{remark}

\usepackage[textsize=tiny]{todonotes}

\usepackage{amsmath,amsfonts,bm}

\def\eqref#1{equation~\ref{#1}}

\def\1{\bm{1}}

\DeclareMathAlphabet{\mathsfit}{\encodingdefault}{\sfdefault}{m}{sl}
\SetMathAlphabet{\mathsfit}{bold}{\encodingdefault}{\sfdefault}{bx}{n}

\newcommand{\E}{\mathbb{E}}
\newcommand{\Ls}{\mathcal{L}}

\newcommand{\Var}{\mathrm{Var}}

\DeclareMathOperator*{\argmax}{arg\,max}

\newcommand{\norm}[1]{\left\lVert#1\right\rVert}

\usepackage{tikz}
\usetikzlibrary{arrows.meta, calc, quotes, tikzmark}
\usetikzlibrary{positioning}

\icmltitlerunning{Which Model to Trust}

\begin{document}

\twocolumn[
\icmltitle{Which Model to Trust: \\Assessing the Influence of Models on the Performance of Reinforcement Learning Algorithms for Continuous Control Tasks}



\icmlsetsymbol{equal}{*}

\begin{icmlauthorlist}
\icmlauthor{Giacomo Arcieri}{eth}
\icmlauthor{David W{\"o}lfle}{fzi}
\icmlauthor{Eleni Chatzi}{eth}
\end{icmlauthorlist}

\icmlaffiliation{eth}{Institute of Structural Engineering, ETH Z{\"u}rich, Z{\"u}rich, Switzerland}
\icmlaffiliation{fzi}{Intelligent Systems and Production Engineering, FZI Research Center for Information Technology, Karlsruhe, Germany}

\icmlcorrespondingauthor{Giacomo Arcieri}{giacomo.arcieri@ibk.baug.ethz.ch}

\icmlkeywords{Reinforcement Learning, model-based Reinforcement Learning, Bayesian Neural Networks, Gaussian Processes, continuous control, benchmark}

\vskip 0.3in
]



\printAffiliationsAndNotice{}  

\begin{abstract}
  The need for algorithms able to solve Reinforcement Learning (RL) problems with few trials has motivated the advent of model-based RL methods. The reported performance of model-based algorithms has dramatically increased within recent years. However, it is not clear how much of the recent progress is due to improved algorithms or due to improved models. While different modeling options are available to choose from when applying a model-based approach, the distinguishing traits and particular strengths of different models are not clear. The main contribution of this work lies precisely in assessing the model influence on the performance of RL algorithms. A set of commonly adopted models is established for the purpose of model comparison. These include Neural Networks (NNs), ensembles of NNs, two different approximations of Bayesian NNs (BNNs), that is, the Concrete Dropout NN and the Anchored Ensembling, and Gaussian Processes (GPs). The model comparison is evaluated on a suite of continuous control benchmarking tasks. Our results reveal that significant differences in model performance do exist. The Concrete Dropout NN reports persistently superior performance. We summarize these differences for the benefit of the modeler and suggest that the model choice is tailored to the standards required by each specific application.
\end{abstract}

\section{Introduction}

Reinforcement Learning (RL) aims to learn an optimal strategy in a sequential decision-making setting through the interactions between an agent and the environment \citep{li2017deep}. 
In recent years, Deep RL has been proven capable of handling high-dimensional problems, some of which considered intractable before, catering to dramatically accelerated progress in the domain of data-driven policy planning \citep{arulkumaran2017deep, mnih2015human, silver2017mastering}.
Nevertheless, the large amount of training data that is required has resulted in limited adoption in real-life applications \citep{polydoros2017survey}.
As a consequence, Model-Based Reinforcement Learning (MBRL) has gained increased attention as this approach is expected to close the gap between simulated and real world tasks \citep{nagabandi2020deep}, due to the higher sample-efficiency compared to model-free algorithms \citep{chua2018deep, gal2016improving, janner2019trust, nagabandi2018neural, wang2019benchmarking}.

MBRL is characterized by the utilization of a model to learn the dynamics of the environment, which is exploited by an RL algorithm in order to solve the underlying planning task. 
Considering recent publications in the field of MBRL, one common choice is to use Gaussian Processes (GPs) as models, like e.g. in \citet{deisenroth2011pilco}.
Neural Networks (NNs) are a popular alternative with several implementation variants, e.g. deterministic NNs \citep{nagabandi2018neural}, Bayesian NNs relying on the Monte Carlo (MC) Dropout approximation \citep{gal2016improving, gal2016dropout} or ensembles of probabilistic NNs \citep{janner2019trust,chua2018deep}.
However, while each of these publications has claimed improved performance on the evaluated environments, it is rarely seen that the model choice is motivated in detail or that several models are compared at all.
As furthermore the RL parts of the proposed solutions usually significantly differ, it remains unclear whether claimed progress has been enabled by advances in the RL algorithm or the model. 
While the importance of selecting a suitable model for the performance of an MBRL algorithm is certainly obvious, the choice of the model remains delicate based on the currently existing literature. 
It is therefore evident that a benchmark is needed for providing clarity on the matter.

The main contribution of this work is to establish such a benchmark that evaluates the influence of model choice on MBRL performance.
To this end we begin by selecting a set of models that are commonly utilized in MBRL, i.e. Neural Networks, ensembles of Neural Networks, Bayesian Neural Networks, and Gaussian Processes, as presented in Section~\ref{sec:models}. 
We furthermore contribute a novel benchmark procedure (Section~\ref{sec:algorithm}) that is intended to yield performance metrics comparable to common MBRL research while still offering a fair comparison of the selected models.
Based on this approach we present benchmark results in Section~\ref{sec:evaluation} for a suite of six different continuous control  environments of increasing complexity that are commonly utilized for the performance evaluation of RL algorithms.
Since sample-efficiency is the main reason to choose model-based algorithms, it is particularly assessed how fast "good" performance is reached for each combination of model and environment.
Besides, overall performance, model robustness, and computing time are evaluated as well.
Finally, Section \ref{sec:comparison} discusses the relationship between prior work and this paper, in particular it is demonstrated that the results of this work can be expected to be more expressive for MBRL tasks in general than existing studies.

\section{Background}
In RL problems an agent observes the environment's state $s_t\in\mathcal{S}$, it selects an action $a_t\in\mathcal{A}$ and receives the reward $r_t$ for the decision made. Next, the environment transitions to the next state $s_{t+1}\in\mathcal{S}$ and a new step of the sequential decision making problem initiates. The goal is to learn a policy, i.e. a sequence of actions over a prescribed horizon, that maximizes the sum of future rewards \citep{sutton2018reinforcement}.

Formally, a RL problem can be stated as a Markov Decision Process (MDP) \citep{van2012reinforcement}. In this work, we consider RL problems that can be described as fully observable MDPs defined by the tuple $(\mathcal{S}, \mathcal{A}, f, r, H)$. Here, $\mathcal{S}$ and $\mathcal{A}$ are the state and the action space, respectively, which are both assumed continuous. Function $f:\mathcal{S} \times \mathcal{A} \rightarrow\mathcal{S}$ is the deterministic transition dynamics $f(s_t,a_t)=s_{t+1}$ of the environment, while $r:\mathcal{S} \times \mathcal{A} \rightarrow\mathcal{R}$ represents the reward function $r(s_t,a_t)=r_t$. In MBRL, this is usually assumed known \citep{nagabandi2018neural, wang2019benchmarking}. Finally, $H$ is the horizon of the problem: the considered MDPs are episodic, namely the state is reset after each episode of lenght $H$.

In MBRL, the interactions with the environment are collected in a dataset $\mathcal{D}=\{(s_t,a_t,s_{t+1})\}$. The agent thus learns a model of the transition dynamics of the environment $\hat{f}(s_t,a_t)=s_{t+1}$. This is exploited to simulate the environment and predict the next states for a sequence of actions. Hence, the learned model can be used for planning the optimal sequence of actions to be executed in this environment. In order to define such a policy, techniques such as stochastic or trajectory optimization are commonly applied.

Finally, central to MBRL is the concept of uncertainty. Following \citet{kendall2017uncertainties} we distinguish two kinds of uncertainty. Aleatoric or statistical uncertainty is the variance produced by some noise intrinsic in data. Epistemic or model uncertainty is the uncertainty over model parameters and depends on the limited amount of training data available. Modeling the epistemic uncertainty is crucial in MBRL to cope with an imperfect model and know when ``the model does not know'' \citep{kendall2017uncertainties}. In addition, it can also drive agent exploration (see Section \ref{sec:exploration}).

\section{Models}\label{sec:models}
This section introduces the models utilized for our evaluation and the motivations that make them interesting modeling choices within MBRL. Furthermore, these models have already been applied successfully to MBRL problems \citep{nagabandi2018neural, chua2018deep, clavera2018model, Gal2017, pearce2018bayesian, pearce2020uncertainty, deisenroth2011pilco, deisenroth2013gaussian}.

\subsection{Deterministic NNs}
 NNs are powerful models, which led to the rise of Deep Reinforcement Learning \citep{arulkumaran2017deep}. We refer to this model as fully deterministic: given the model parameters, there is a unique prediction for a specific input. Besides, this model is not able to recover the full distribution that generated the observations nor to model any kind of uncertainty \citep{chua2018deep}. As all other NN-based methods, it is also referred to as a model of finite capacity \citep{gal2016improving}, which needs to smooth over all data points. Moreover, NN-based models take several training steps for learning, in contrast to GPs. On the other hand, NNs scale very well with dimensionality and are able to handle complex tasks. The Deterministic NN is also the model adopted by the model-based Deep RL pioneering work in \citet{nagabandi2018neural} for learning the dynamics of the environment. While many advances have been produced in recent years over this model, some of which are also introduced in the following sections, it is still meaningful to select it for our comparison to draw a baseline and understand whether these constitute actual improvements.

\subsection{Deterministic Ensembles}
We define the Deterministic Ensemble as an ensemble of several deterministic NNs. It is well-known that by aggregating a group of predictors, this leads in improved and more robust performance than any single predictor alone \citep{zhou2019ensemble}. Further, an ensemble of NNs extends the previously described model by providing an estimate of model uncertainty: the variance of the networks' predictions is an approximate measure of the epistemic uncertainty \citep{beluch2018power}. This technique has already demonstrated interesting results \citep{lakshminarayanan2016simple}. While it is important that the NNs are mutually independent, it was empirically proved that bootstrapping is unnecessary, and the stochasticity of the optimizer and the random initialization of the weights make the NNs sufficiently independent \citep{lee2015m}. However, the NNs that compose the ensemble are fully deterministic models and do not present any inherent regularization (such as a prior distribution over the weights). As a result, the NNs may converge to similar parameters, nullifying the benefits of the ensemble. Furthermore, deterministic NNs are also more prone to overfit data compared to Bayesian models \citep{blundell2015weight}.

\subsection{Concrete Dropout NNs}
Concrete Dropout \citep{Gal2017} is a practical approximation for applying the framework of Bayesian deep learning, which promises improved robustness and superior capability for well-calibrated uncertainty estimates. In general, Bayesian NNs are probabilistic models which replace the deterministic parameter values of classical NNs with entire parameter distributions over the weights \citep{bnnMackay}. Given a prior distribution $p(\theta)$ and some data $D=(\mathbf{x}, \mathbf{y})$, the posterior distribution $p(\theta|D)$ is computed via Bayes rule. Given new input data $\mathbf{x_{new}}$, it is thus possible to recover the posterior predictive distribution by marginalizing over all posterior parameters:
\begin{equation}
    p(\mathbf{y}|\mathbf{x_{new}}, D)=\int_{\theta}p(\mathbf{y}|\mathbf{x_{new}}, \theta)p(\theta|D)
\end{equation}
While it is possible to apply this framework exactly for small-scale problems, computing $p(\theta|D)$ and $p(\mathbf{y}|\mathbf{x_{new}}, D)$ is generally analytically intractable due to the large number of parameters and to the nonlinear functional form of a NN \citep{blundell2015weight}. If we assume $p(\theta|D)$ known, the posterior predictive distribution is empirically recovered with MC approximation. Let $\mathbf{f}^{\hat{\theta}}(\mathbf{x}_i)$ be the output of the BNN for a sample $\mathbf{\hat{\theta}}\sim p(\theta|D)$
and a given input $\mathbf{x}_i$. If one repeats the prediction $T$ times, every time a different sample from the posterior weight distribution is obtained, which yields different outputs. That
is, we are taking samples from the posterior predictive distribution $\mathbf{f}^{\hat{\theta}}(\mathbf{x}_i)\sim p(\mathbf{y}|\mathbf{x_{i}}, D)$.
Provided T is large enough, we can approximate the predictive mean as follows:
\begin{equation}\label{eq:mc_mean}
    \E(\mathbf{y}) \approx \frac{1}{T}\sum^T_{t=1}\mathbf{f}^{\hat{\theta}}(\mathbf{x}_i)
\end{equation}
Unfortunately, recovering $p(\theta|D)$ is an extremely challenging task and the need for frameworks that may sufficiently approximate the posterior weight distribution has long limited the use of BNNs. Concrete Dropout forms one of the two solutions we employ in this work.

Dropout is a technique widely adopted in deep learning for enhancing generalization and improving performance in out-of-sample predictions \citep{hinton2012improving}. At each training step, every parameter has a probability $p$ to be turned off and a probability $1-p$ to be turned on. As a result, a different NN configuration is sampled at each training step. \citet{gal2016dropout} have shown that turning on dropout at testing time (MC dropout) can be interpreted as a variational inference approximation of BNNs with a Bernoulli variational distribution. While MC dropout gained large popularity in supervised learning tasks, two reasons inhibited its application in RL. First, the technique requires hand tuning over the dropout probabilities, with prohibitive computing time in RL applications. Second, the dropout probability should be adapted to the data at hand, making this operation also infeasible within RL where the amount of data changes over time. Ideally, the dropout rate should be higher at the beginning of the RL training and it should decrease as the amount of data increases. To this end, \citet{Gal2017} conceived the Concrete Dropout variant tailored for RL applications. This can be seen as a continuous relaxation of the discrete dropout technique. Consequently, the Concrete Dropout is able to use the derivative estimator to self-optimize the dropout rate for each layer and at each training step, solving both the issues. The Concrete Dropout matches the performance of hand-tuned MC
dropout on several tasks and allows the use of BNNs with dropout in RL applications. Following \citet{Gal2017} and \citet{kendall2017uncertainties}, the NN output is decomposed into:
\begin{equation}
    \mathbf{f}^{\hat{\theta}}(\mathbf{x})=(\mathbf{\hat{y}},\hat{\sigma}) 
\end{equation}
The predictive mean distribution is then computed as in Equation \ref{eq:mc_mean}, while the predictive variance is approximated as:
\begin{equation}
    \Var(\mathbf{y})\approx\left[\frac{1}{T}\sum_{i=1}^T\mathbf{\hat{y}}_i^2-\left(\frac{1}{T}\sum_{i=1}^T\mathbf{\hat{y}}_i\right)^2\right]+\frac{1}{T}\sum_{i=1}^T\hat{\sigma}_i^2
\end{equation}
The first term is the aforementioned epistemic uncertainty and it vanishes when $\hat{\theta}_i$ is a constant for all $i$, i.e. when we have zero parameter uncertainty. The second term represents the aleatoric uncertainty, namely the intrinsic noise in data.

\subsection{Anchored Ensembling} 
Anchored Ensembling \citep{pearce2020uncertainty} represents the second candidate solution we adopt for applying BNNs in practice. The technique approximates Bayesian inference through an ensemble of NNs with parameters regularised around values drawn from an anchor distribution. This exploits a procedure which falls into a family of Bayesian methods called Randomised MAP Sampling (RMS). The RMS procedure follows the premise that addition of a regularisation term to a loss function returns a MAP parameter estimate, namely a point estimate of the Bayesian posterior. Injecting noise into this loss and sampling repeatedly (i.e. ensembling) results in a distribution of MAP solutions approximating the true posterior. \citet{pearce2020uncertainty} prove that under two conditions, partly present in NNs, their approximate method samples from the true posterior distribution. The authors show that their method is able to estimate an epistemic uncertainty extremely close to the one of a Gaussian Process (considered the ground truth). We particularly choose to employ this model for the expected superior capacity in providing well-calibrated uncertainty estimates. Furthermore, this model achieved competitive performance with other methods in supervised learning tasks and it has already been used for RL applications \citep{pearce2018bayesian}.

\subsection{Gaussian Processes} 
GPs \citep{williams2006gaussian} are probabilistic non-parametric models widely used for applying Bayesian inference in regression tasks. GPs can be defined as an infinite set of random variables, completely defined by a mean function $\mu$ and a covariance function (also called kernel) $k$: given $x, x' \in X$, $p(x) \sim \mathcal{GP}(\mu(x), k(x, x'))$. It is thus a distribution over functions. Function $\mu$ is often assumed equal to zero, since more complicated mean functions do not improve
substantially the performance. When the GP has not yet observed the training data, the prior distribution lies
around $\mu=0$, according to the previous assumption. When the GP receives the information, the covariance matrix determines the functions, from the space of all possible functions, which are more probable. The conditioning property constrains this set of functions to pass through each training data. The GP model assigns a random variable to each point to predict and, through the marginalisation of each random variable, it computes the mean function value $\mu'_i$ and the standard deviation $\sigma'_i$ of the $i$-th test point, where $\mu'_i$ is no longer equal to zero. If a test point is located on the training data, the function goes directly through it. Otherwise, the predictions are influenced by the training data proportionally to their distance, according to
the kernel. 

This model has been selected as it represents the principal counterpart of NNs for learning the dynamics of the environment in MBRL \citep{deisenroth2011pilco, deisenroth2013gaussian}. Since BNNs of infinite width are proved to converge to GPs \citep{neal2012bayesian}, they are also considered the ground truth to evaluate the epistemic uncertainty estimation of approximate models. Moreover, its infinite capacity makes this model able to learn and memorize all observed data without the need of different training steps, in
contrast to NN models \citep{gal2016improving}. On the other hand, this model is known to scale cubically with number of data points and it is also negatively affected by problem dimensionality. 

\section{Benchmark Procedure}\label{sec:algorithm}
This section describes the algorithm employed for the model comparison benchmark and it presents all implementation details of the considered models. A pure MBRL algorithm is usually composed of the general controller, the optimization algorithm, and the exploration policy. The choice of the MBRL algorithm is crucial for the success of the work. It should be tailored to serve our model comparison purpose, highlighting how well each model has learned the transition dynamics, while still delivering good performance. Considering this trade-off is key.

As a general controller, we employ the widely adopted \textit{Model Predictive Control} (MPC) \citep{nagabandi2018neural, chua2018deep, wang2019benchmarking}, which is reported in Algorithm \ref{alg:mpc}. The MPC controller is key to reach good performance even with an imperfect dynamics model. Its peculiarity is contained in Line \ref{algl:mpc}, where only the first action from the optimal trajectory is executed. As a result, the agent re-plans the optimal sequence of actions at each time step, allowing to handle unexpected outcomes. Thus, the agent
really interacts with the environment and a closed-loop policy is achieved.

\begin{algorithm}[H]
\caption{Model Predictive Control}
\label{alg:mpc}
\begin{algorithmic}[1]
\STATE Run random policy $\pi_0$  to collect data $\mathcal{D}=\{(\mathbf{s}_t, \mathbf{a}_t, \mathbf{s}_{t+1})_i\}$
\FOR{$\mathbf{K}$ episodes}

  \STATE	Learn the dynamics model $\hat{f}(s_t, a_t)=s_{t+1}$. \label{algl:learn}
  	\FOR{$H$ steps}
  
  	    \STATE Plan through $\hat{f}(s, a)$ to choose actions. \label{algl:opt}
  		\STATE Execute the first planned action only and observe the resulting state $s_{t+1}$ \label{algl:mpc}
  		\STATE Append $(\mathbf{s}_t, \mathbf{a}_t, \mathbf{s}_{t+1})$ to $\mathcal{D}$
  	\ENDFOR
\ENDFOR
\end{algorithmic}
\end{algorithm}

\subsection{Learning the Dynamics Model}
All NN-based models are composed by feed-forward fully-connected stacked layers and ReLu hidden activation functions. The selected hyperparameters are reported in Appendix \ref{app:hidden_size}. The Deterministic NN and the Deterministic Ensemble are trained with an MSE loss function:
\begin{equation}
    \Ls\textsubscript{MSE}(\hat{\theta}, \mathbf{x},\mathbf{y})=\frac{1}{N}\sum_{i=1}^N\norm{\mathbf{y}_i-\mathbf{\hat{y}}_i}^2
\end{equation}
As in \citet{kendall2017uncertainties}, we assume a Gaussian likelihood and train the Concrete Dropout NN with the heteroscedastic loss:
\begin{equation}
    \Ls\textsubscript{h}(\hat{\theta}, \mathbf{x}, \mathbf{y}) = \frac{1}{N}\sum_{i=1}^N\frac{1}{\hat{\sigma}_i^2}\norm{\mathbf{y}_i-\mathbf{\hat{y}}_i}^2+\log\hat{\sigma}_i^2
\end{equation}
The Anchored Ensembling is instead trained with the regularised MSE loss function, as in \citet{pearce2020uncertainty}:
\begin{equation}
    \Ls\textsubscript{anc,}_j(\hat{\theta}_j, \mathbf{x}, \mathbf{y}) =  
    \Ls\textsubscript{MSE}(\hat{\theta}_j, \mathbf{x}, \mathbf{y})  + \frac{1}{N}\norm{\Gamma^{1/2}(\hat{\theta}_j-\theta_{anc,j})}^2
\end{equation}
where the subscript $j$ refers to the $j$-th NN of the ensemble, $\Gamma$ is a diagonal regularisation matrix, and $\theta_{anc, j}$ is the initialization drawn from the anchor distribution. For the GP model, the covariance function determines the learning performance. As suggested in \citet{stein1999interpolation} and \citet{williams2006gaussian}, we adopt the \emph{Matèrn52} class.

Following \citet{deisenroth2013gaussian}, the training target $\mathbf{y}$ is not directly the next state $s_{t+1}$, but the difference between the next state and the current state $\Delta_{s_{t}} = s_{t+1} - s_{t}$. The artifice was introduced to prevent GPs from falling to 0 in uncertain zones. This technique turned out to strongly reduce performance variance also in NNs and it has subsequently been adopted apart from GPs \citep{nagabandi2018neural}, and is thus applied to all models considered in this work.
Lastly, all models are trained on a replay buffer \citep{lillicrap2020continuous} which keeps only the last $N=2000$ data and prevent the dataset $\mathcal{D}$ from quickly containing an incredibly large amount of data. This trick also easily solves the issue of giving more importance to new trials than old ones \citep{gal2016improving}. However, a well-known disadvantage is that the model might forget important old episodes.

In order to compute the predictive mean (Equation \ref{eq:mc_mean}), we employ 5 NNs for the two ensembling methods. This is a number generally accepted to achieve satisfying results for both Deterministic \citep{beluch2018power, lakshminarayanan2016simple} and Anchored \citep{pearce2020uncertainty} Ensembles. Concerning the Concrete Dropout NN, we follow the author's implementation \citep{gal_cd_imp} and repeat the MC prediction $T=20$ times.

\subsection{The Optimization Algorithm}\label{sec:opt_alg}

\begin{table*}[!t]
\caption{Benchmark results: models performance are reported with mean returns and standard error over all different tasks. Best results are highlighted in bold.}
\label{tab:results}
\begin{center}
\begin{tabular}{lcccccc}
\multicolumn{1}{c}{\bf Model}  &\multicolumn{1}{c}{\bf Pendulum}  &\multicolumn{1}{c}{\bf InvertedPendulum}  &\multicolumn{1}{c}{\bf InvertedDoublePendulum}
\\ \hline \\
Deterministic NN         & $-424.68\pm25.76$ & $-279.52\pm39.41$ & $-66.19\pm2.08$ \\
Deterministic Ensemble   & $-352.50\pm22.41$ & $\pmb{-\ \ 81.48\pm18.98}$ & $-48.37\pm1.82$ \\
Concrete Dropout NN      & $\pmb{-315.40\pm18.70}$ & $-126.17\pm29.67$ & $\pmb{-34.95\pm1.67}$ \\
Anchored Ensembling      & $-407.99\pm26.62$ & $-289.99\pm40.06$ & $-58.54\pm2.03$ \\
Gaussian Process         & $-417.49\pm24.33$ & $-115.58\pm34.45$ & $-53.03\pm1.98$ 
\\ \\
\multicolumn{1}{c}{\bf Model}  &\multicolumn{1}{c}{\bf Reacher}  &\multicolumn{1}{c}{\bf HalfCheetah}  &\multicolumn{1}{c}{\bf Hopper}
\\ \hline \\
Deterministic NN         & $-35.22\pm0.54$ & $-80.08\pm1.63$ & $-750.75\pm41.27$ \\
Deterministic Ensemble   & $-34.05\pm0.48$ & $-71.32\pm1.46$ & $-601.73\pm38.52$ \\
Concrete Dropout NN      & $\pmb{-31.22\pm0.35}$ & $\pmb{-27.09\pm1.72}$ & $\pmb{-368.25\pm37.50}$ \\
Anchored Ensembling      & $-34.58\pm0.51$ & $-81.67\pm1.49$ & $-715.41\pm38.70$ \\
Gaussian Process         & $-31.80\pm0.36$ & $-78.18\pm1.50$ & $-765.24\pm43.83$ \\
\end{tabular}
\end{center}
\end{table*}

The core of the MBRL algorithm is composed by the online optimization algorithm that plans through the learned model (Line \ref{algl:opt} of MPC). As in \citet{nagabandi2018neural}, we apply the Random Shooting (RS) method, which is reported in Algorithm \ref{alg:RS}.

\begin{algorithm}[H]
\caption{Random Shooting}
\label{alg:RS}
\begin{algorithmic}[1]
  \STATE Set number of trajectories $N$ and planning horizon $n$
  \STATE Pick $\mathbf{A}_1,\dots, \mathbf{A}_N$ from uniform distribution $p(\mathbf{A})$, where $\mathbf{A}_i=\{a_{0,i}, $\dots$, a_{n,i}\}$\\
  \FOR{trajectory $i=1$ {\bfseries to} N}
    
        \FOR{step $t=0$ {\bfseries to} $n-1$}
        
        \STATE Predict $\hat{f}(s_{t,i}, a_{t,i})=s_{t+1,i}$
        \ENDFOR
        \STATE Evaluate the trajectory $J(\mathbf{A}_i)=\sum_{t=0}^nr(s_{t,i}, a_{t,i})$ \label{algl:eval}
  \ENDFOR
  \STATE Select $\mathbf{A}_{best}=\argmax_i J(\mathbf{A}_i)$.
  
\end{algorithmic}
\end{algorithm}

Based on the findings in \citet{nagabandi2018neural} and \citet{wang2019benchmarking}, we set the number of trajectories $N$ to 500 and the planning horizon $n$ to 20. 

RS has been selected as optimization algorithm as the shooting methods family does not make strong assumptions on the transition dynamics and the reward function. Furthermore, RS has recently shown competitive results with more sophisticated algorithms such as Cross-Entropy Method (CEM) when combined with complex models \citep{wang2019benchmarking}. Finally, RS performance heavily depends on model accuracy. As a result, models that have learned better transition dynamics achieve better performance. This makes RS perfect for our model comparison objective and allows us to consider the sum of rewards per episode achieved during the RL training as a natural estimate of model performance. 

As a common practice in MBRL, we also assume to access the reward function of the environment in order to compute Line \ref{algl:eval} of RS. This allows to separate the dynamics model from the task to accomplish. Main benefit is obtaining a general model which can be applied to solve different tasks by only changing the reward function \citep{nagabandi2018neural}.

\subsection{The Exploration Policy}
\label{sec:exploration}

The last piece of the benchmarking algorithm is formed by the exploration policy. This is needed to visit new states and potentially improve the dynamics model. Here, we try to fully exploit the diverse model capacities. Consequently, we implement two different policies depending on whether the employed model is able to estimate the epistemic uncertainty. For the Deterministic NN, a simple $\varepsilon$-greedy policy \citep{azizzadenesheli2018efficient} is used. For all other models, we conceive a simple but effective exploration policy, which maximizes the Information Gain (IG) from visiting new states. This follows from the approximation of the IG with the \textit{prediction gain} \citep{bellemare2016unifying}. Intuitively, a state is more informative if it causes a large change in model parameters. Let $\mathcal{U}(\cdot)$ be the function that outputs the epistemic uncertainty given a state and an action. The IG-based policy works as follows:

\begin{algorithm}[H]
\caption{IG-based exploration policy}
\label{alg:IG}
\begin{algorithmic}[1]
  \STATE Randomly sample $N$ actions $a_1,\dots, a_N$
  \FOR{action $i=1$ {\bfseries to} N}
  
        \STATE Evaluate the epistemic uncertainty $\mathcal{U}(s_t, a_i)$
  \ENDFOR
  \STATE Select $a_{IG}=\argmax_i \mathcal{U}(s_t, a_i)$
  
\end{algorithmic}
\end{algorithm}

The full MBRL algorithm is reported in pseudo-code in Appendix \ref{app:full}.

\section{Results}\label{sec:evaluation}

\begin{figure*}[htb]
\begin{minipage}{0.5\textwidth}
\begin{tikzpicture}
  \node (img)  {\includegraphics[scale=0.3]{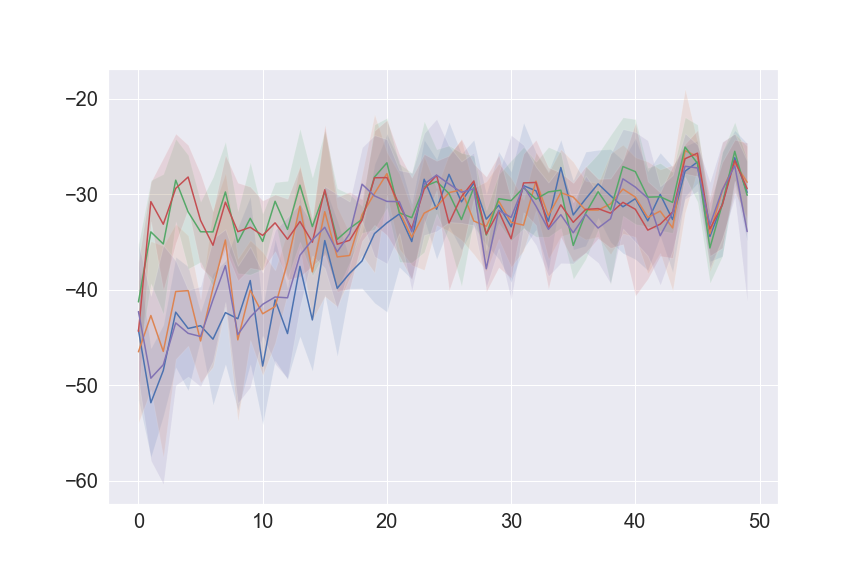}};
  \node[above=of img, node distance=0cm, yshift=-1.7cm] {Reacher};
  \node[below=of img, node distance=0cm, yshift=1.5cm] {Episode};
  \node[left=of img, node distance=0cm, rotate=90, anchor=center,yshift=-1.3cm] {Sum of rewards};
 \end{tikzpicture}
\end{minipage}%
\begin{minipage}{0.5\textwidth}
\begin{tikzpicture}
  \node (img)  {\includegraphics[scale=0.3]{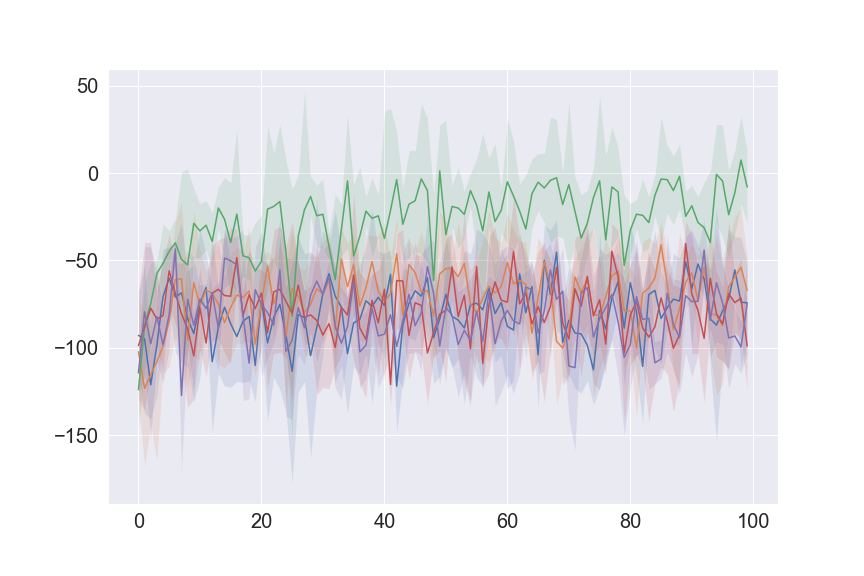}};
  \node[above=of img, node distance=0cm, yshift=-1.7cm] {HalfCheetah};
  \node[below=of img, node distance=0cm, yshift=1.5cm] {Episode};
  \node[left=of img, node distance=0cm, rotate=90, anchor=center,yshift=-1.3cm] {Sum of rewards};
\end{tikzpicture}
\end{minipage}%
\caption{Model performance on Reacher (left) and HalfCheetah (right) environments. Deterministic NN (blue), Deterministic Ensemble (orange), Concrete Dropout NN (green), Gaussian Process (red), and Anchored Ensembling (purple) are displayed.}
\label{fig:rh}
\end{figure*}
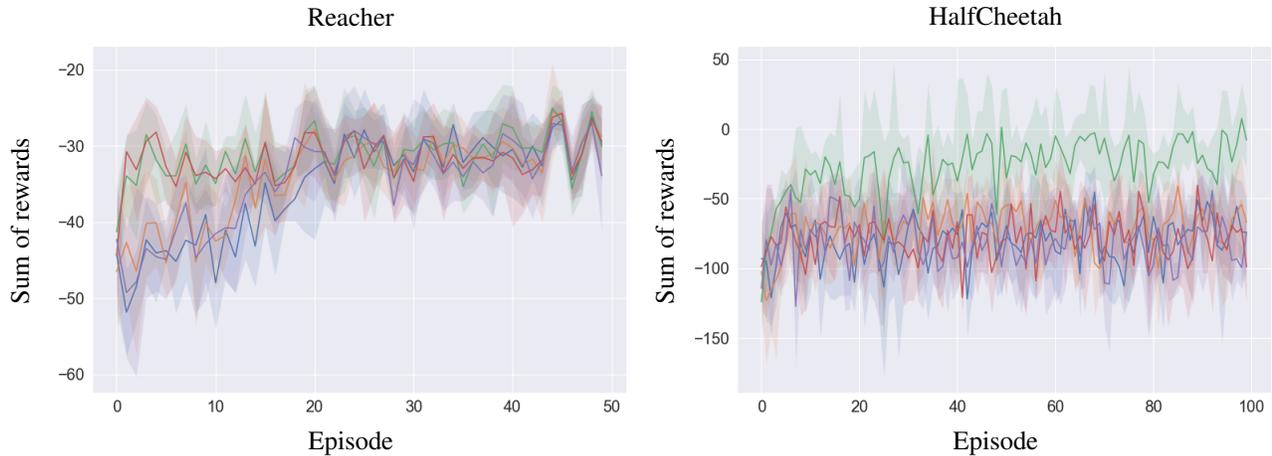

This section presents the benchmark results, i.e., the MBRL performance of the considered models for the selected environments. The latter are composed by one OpenAI Gym \citep{brockman2016openai} and five PyBullet Gym \citep{benelot2018} environments, listed in Appendix \ref{app:env}. In order to access the reward functions (see Section \ref{sec:opt_alg}), these are implemented as reported in Appendix \ref{app:reward}. The employed model configurations are outlined in Appendix \ref{app:hidden_size} for all tasks to ease reproducibility and transparency of the results.

The performance for each model is summarized in terms of average returns and standard error over 5 different random seeds. Table \ref{tab:results} depicts the benchmarks results.

The Concrete Dropout NN delivers the best performance in 5 out of 6 environments. Especially in the most complex tasks, such as HalfCheetah and Hopper, it dramatically outperforms all other models. The results further prove the high robustness provided by the dropout technique as a Bayesian approximation. In addition, this is the only model able to learn both the aleatoric and the epistemic uncertainty. This supports the intuition in \citet{lakshminarayanan2016simple} that learning both the two types of uncertainty leads to improved performance.  

The Deterministic Ensemble takes second place in most of the environments and even delivers the highest performance in the InvertedPendulum problem. With significantly higher results than the Deterministic NN, it proves that ensembling does improve performance in MBRL. 

The Gaussian Process delivers high performance in Reacher, which is a relatively complex environment, and in easier tasks such as InvertedPendulum. However, as the complexity of the task increases, it is no longer possible to appreciate satisfying results. Indeed, it completely fails in HalfCheetah and Hopper. It also reports strong instability across different random seeds. Nevertheless, we recognize that the poor performance may be due to the use of the fixed replay buffer. 

Unexpectedly, the Anchored Ensembling does not show substantially better performance when compared to the Deterministic NN. Along with the GP, it is affected by high instability across different random seeds.

Deepening our findings, we also show in  Figure \ref{fig:rh} two representative plots, whereby model performance is depicted by ``Average Return'' and ``Standard Deviation of Returns'', as suggested in \citet{islam2017reproducibility}, in Reacher and HalfCheetah environments. No smoothing technique is applied to the results shown. Learning curves denote the mean performance, while the shaded regions represent one standard deviation of rewards over 5 different random seeeds. In the Reacher task, the Concrete Dropout NN and the Gaussian Process show an impressive sample-efficiency. While these two models converge to good results after only a couple of episodes, the others need about 20 episodes to achieve similar performance. In HalfCheetah, the Concrete Dropout NN clearly succeeds in terms of both sample-efficiency and overall performance, delivering unattainable results for all other models.

Lastly, it is noteworthy to consider the computing time of different models over the benchmarking comparison. As one might expect, the Deterministic NN is the fastest model across all tasks. Deterministic and Anchored Ensembles have a constant ratio with about 5 times longer computing time than the previous model. The Concrete Dropout NN and the Gaussian Process starts at a similar ratio than the former two in the easier tasks, but decay quickly as the complexity increases. In HalfCheetah, the computing times become about 15 and 30 times more than the Deterministic NN, respectively.

\section{Comparison to Related Work}
\label{sec:comparison}

\citet{chua2018deep} present one of scarce - if not the only - attempt to compare several models while proposing a MBRL algorithm.
Regarding model selection the main finding of this work is that models capable of reflecting aleatoric and epistemic uncertainty, defined as a probabilistic ensemble, likely lead to better performance on complex tasks.
However, the study does not aim to provide a benchmark to assess model influence on MBRL, nor does it consider alternative models capable of capturing both types of uncertainty, as our work does.

\citet{wang2019benchmarking} recognized the difficulty in offering an objective comparison between available MBRL methods.
In order to quantify scientific progress, the authors proposed the first benchmark in MBRL.
However, their work substantially differs from ours as it focuses on comparing the MBRL algorithms as a combination of models and planning strategies, whereas we focus on the single aspect of model class selection.
While the study is certainly valuable for the scientific progress in MBRL, it provides no guidance for model choice as our work does.

Finally, \citet{kegl2021model} is the only prior work with a similar intention, i.e. comparing different models for model-based RL algorithms.
However, this work significantly differs from ours in two points.
Firstly and as acknowledged by the authors themselves, the choice to evaluate the models on a single environment (Acrobot, a relatively low-dimensional discrete control task) does not provide strong evidence for the generalization of the results to other tasks. In contrast we evaluate on a larger suite of continuous control tasks and thus expect better generalization of our results.
Secondly, the benchmark in \citet{kegl2021model} focuses on autoregressive and multimodal models, without considering popular MBRL approaches such as BNNs as our work.
This difference is particular relevant as their results highlight the superiority of Mixture Density Networks \citep{bishop1994mixture}, even though the employed environment is well known to be particularly well suited for multimodal models.

\section{Conclusion}\label{sec:conclusion}

This work investigates the influence of models on the performance of RL algorithms. Model-based methods are expected to solve RL problems with fewer trials
(i.e. higher sample-efficiency). However, the choice of model within the RL architecture has not formally been investigated in terms of its effect in improving the performance reported by novel algorithms. This work addresses this issue by comparing typical classes that are commonly used as models for MBRL. The selected models are deterministic Neural Networks, ensembles of deterministic Neural Networks, Concrete Dropout Neural Networks, Anchored Ensembles, and Gaussian Processes. The model comparison benchmark is carried out  on six environments of increasing complexity.

Our results show that significant differences in model performance do exist. The Concrete Dropout NN reports  persistently superior performance on 5 out of 6 environments. Especially in the most complex tasks, it dramatically outperforms all other models. Its distinguishing strengths are higher sample-efficiency, robustness, and flexibility over different task complexities. On the negative side, this model causes higher computational load than other NN-based models. Without the need of different training steps to memorize all data, main strength of the GP is the remarkable sample-efficiency within the first handful of trials in low or medium dimensional problems. However, its lack of consistency prevents this model from converging to high performance compared to other explored models. As task complexity increases, it requires dramatically higher computing time and it is no longer possible to appreciate satisfying results. While the Deterministic Ensemble consistently achieves higher returns than the Deterministic NN, the differences are reduced in high dimensional problems. This suggests a lower flexibility in terms of task complexity when compared to the Concrete Dropout NN. The Anchored Ensembling is not generally able to outperform the Deterministic NN across the evaluated tasks. Lastly, the Deterministic NN does not deliver remarkable performance in any task but it succeeds in terms of computing time. 

Our work proves that it is important that differences among models are taken into account. The improved performance reported by novel algorithms may (partly) be produced by the choice of different modeling approaches. Performance of novel MBRL algorithms, which introduce both a new RL approach and a different modeling choice, should be compared not only against previous methods, but also formally investigated with different modeling options in a similar fashion that we introduced in this work. Innovative techniques should therefore distinguish and make clear how much of the enhanced performance is due to improved algorithms or due to improved models.

Finally, our results indicate that model choice can be tailored to the application of the RL algorithm. In the case where there are no computing time requirements, the Concrete Dropout NN is consistently the favorite candidate among those explored to deliver the highest sample-efficiency and overall performance, regardless of task complexity. However, if the task to accomplish is relatively low-dimensional and has to be solved only once with a few samples, also Gaussian Processes represent a valid alternative. On the other hand, if decisions must be made in the shortest possible time, a simpler Deterministic NN may be preferred.


\bibliography{bibliography}
\bibliographystyle{icml2022}

\newpage
\appendix
\onecolumn

\section{Algorithm}\label{app:full}
\begin{algorithm}[H]
\caption{The MBRL algorithm}
\begin{algorithmic}[1]
  
  \STATE {\bfseries Input:} Number of episodes $\mathbf{K}$, episode length $H$, $\varepsilon$ probability to explore, number of actions sampled for IG exploration, number of trajectories and planning horizon $n$ for Random Shooting.
  \STATE {\bfseries Output:} Sum of rewards list $\mathbf{R}=\{R_0,\dots,R_{\mathbf{K}}\}$, with $R_i=\sum_{t=0}^Hr_t$.
  \STATE {\bfseries Data:} $\mathcal{D}=\{\}$ replay buffer with capacity 2000 steps.
  \STATE Initialize $\mathcal{D}$ with a random controller for one episode.
  \FOR{episode = 1 to $\mathbf{K}$}
  
  		\STATE Train dynamics model $\hat{f}$ on $\mathcal{D}$.
  		\FOR{step t = 0 to H}
  	
  			\IF{$p\sim U(0,1)<\varepsilon$}
  			
  				\STATE Do exploration to find $a_{expl}$. 	\COMMENT{IG exploration or random exploration depending on the model}
  				\STATE $a_t:=a_{expl}$
  			\ELSE
  			
  				\STATE Do Random Shooting to select $\mathbf{A}_{best}=\{a_0,\dots,a_n\}$.
  				\STATE $a_t:=a_0$
  			\ENDIF
  			\STATE Execute $a_t$ and collect reward $r_t$.
  			\STATE Update dataset $\mathcal{D}\leftarrow\mathcal{D}\cup\{(s_t, a_t, s_{t+1})\}$ 
  	\ENDFOR
  \ENDFOR
\end{algorithmic}
\end{algorithm}

\section{Environments}\label{app:env}
\smallskip
\begin{table}[!ht]
\caption{Dimensionality, horizon, and number of episodes run of the environments.}
\label{tab:dim_env}
\begin{center}
\begin{tabular}{lcccc}
\multicolumn{1}{c}{\bf Environment}  &\multicolumn{1}{c}{\bf Observation Space} &\multicolumn{1}{c}{\bf Action Space} &\multicolumn{1}{c}{\bf Horizon}
&\multicolumn{1}{c}{\bf Episodes}
\\ \hline \\
Pendulum & 3 & 1 & 200 & 50 \\
InvertedPendulum & 4 & 1 & 100 & 35 \\
InvertedDoublePendulum & 11 & 1 & 100 & 100 \\
ReacherPyBullet & 9 & 2 & 50 & 50 \\
Hopper & 11 & 3 & 500 & 100 \\
HalfCheetah & 17 & 6 & 500 & 100
\end{tabular}
\end{center}
\end{table}

\subsection{Reward Functions}\label{app:reward}

\begin{table}[!ht]
\caption{Reward functions of the environments.}
\label{tab:reward}
\begin{center}
\begin{tabular}{lc}
\multicolumn{1}{c}{\bf Environment}  &\multicolumn{1}{c}{\bf Reward function}
\\ \hline \\
Pendulum & $-(((\theta_t+\pi) \% (2\pi)) - \pi)^2-0.1\dot{\theta}_t^2-0.001a_t^2$ \\
InvertedPendulum & $-\theta_t^2$ \\
InvertedDoublePendulum & $-\theta_t^2-\gamma_t^2-0.001\dot{\theta}_t^2-0.005\dot{\gamma}_t^2
$ \\
Reacher & $-distance_t - \|\mathbf{a}_t\|^2$ \\
Hopper & $\dot{x}_t-0.1\|\mathbf{a}_t\|^2-3h_t^2+1$ \\
HalfCheetah & $\dot{x}_t-0.1\|\mathbf{a}_t\|^2$
\end{tabular}
\end{center}
\end{table}

\newpage
\section{Hidden Size}\label{app:hidden_size}

\begin{table}[!ht]
\caption{Hidden size of NN-based models.}
\label{tab:hidden_size}
\begin{center}
\begin{tabular}{lcccccc}
\multicolumn{1}{c}{\bf Model}  &\multicolumn{1}{c}{\bf Pendulum}  &\multicolumn{1}{c}{\bf InvPendulum}  &\multicolumn{1}{c}{\bf InvDoublePendulum}
\\ \hline \\
Deterministic NN         & $2\times32$ & $2\times40$ & $2\times100$ \\
Deterministic Ensemble   & $2\times32$ & $2\times40$ & $2\times100$ \\
Concrete Dropout NN      & $2\times100$ & $2\times100$ & $2\times200$ \\
Anchored Ensembling      & $2\times40$ & $2\times40$ & $2\times100$ 
\\ \\
\multicolumn{1}{c}{\bf Model}  &\multicolumn{1}{c}{\bf Reacher}  &\multicolumn{1}{c}{\bf HalfCheetah}  &\multicolumn{1}{c}{\bf Hopper}
\\ \hline \\
Deterministic NN         & $2\times200$ & $3\times250$ & $3\times250$ \\
Deterministic Ensemble   & $2\times200$ & $3\times250$ & $3\times250$ \\
Concrete Dropout NN      & $2\times500$ & $3\times1024$ & $3\times1024$ \\
Anchored Ensembling      & $2\times200$ & $2\times500$ & $2\times500$ 
\end{tabular}
\end{center}
\end{table}

\end{document}